# Towards a High-Performance Object Detector: Insights from Drone Detection Using ViT and CNN-based Deep Learning Models

Junyang Zhang

Department of Electrical Engineering and Computer Science & Department of Computer Science University of California, Irvine, CA, United States, 92617

junyanz9@uci.edu

*Abstract:* Accurate drone detection is strongly desired in drone collision avoidance, drone defense and autonomous Unmanned Aerial Vehicle (UAV) self-landing. With the recent emergence of the Vision Transformer (ViT), this critical task is reassessed in this paper using a UAV dataset composed of 1359 drone photos. We construct various CNN and ViT-based models, demonstrating that for single-drone detection, a basic ViT can achieve performance 4.6 times more robust than our best CNN-based transfer learning models. By implementing the state-of-the-art You Only Look Once (YOLO v7, 200 epochs) and the experimental ViT-based You Only Look At One Sequence (YOLOS, 20 epochs) in multi-drone detection, we attain impressive 98% and 96% mAP values, respectively. We find that ViT outperforms CNN at the same epoch, but also requires more training data, computational power, and sophisticated, performance-oriented designs to fully surpass the capabilities of cutting-edge CNN detectors. We summarize the distinct characteristics of ViT and CNN models to aid future researchers in developing more efficient deep learning models.

## I. Introduction

In the commercial drone market, computer vision is the most widely used technology in obstacle avoidance systems due to the physical limitations of infrared ray and ultrasonic wave technologies, as well as the high cost of laser-based obstacle avoidance systems. Both infrared and ultrasonic obstacle avoidance systems have strict requirements concerning the reflecting object and surrounding environment, making these technologies less reliable. For example, infrared light can be absorbed by black objects and penetrate transparent objects, and its receiver can be disturbed by other sources of infrared light. Similarly, ultrasonic waves can be absorbed by sponges and disturbed by propellor airflow. Moreover, drones, known for their speed, compact size, and difficulty to locate and intercept, present unique challenges. During the war in Ukraine, soldiers were injured and killed daily by bomber drones and suicide drones. In current autonomous UAV self-landing technology, GPS is used to locate the drones. However, most GPS systems have around a 1-meter error range, which is not safe enough for a large number of drones to autonomously land. These scenarios highlight the need for a high-performance drone detector, and computer vision presents the most economical and generalizable solution.

Convolutional neural networks (CNNs) have made significant advancements in several areas within the field of computer vision. Typically, contemporary detectors utilize pure convolution networks to draw out features. Traditional image classification networks like VGG 16 [9] and ResNet 50 [5] are used as the foundational architecture for our single object detectors. In the case of the YOLO series of detectors, they utilize a unique residual network known as Darknet, which offers superior efficiency in feature extraction [8]. In this paper, we deploy YOLO v7 [11] on the multiple drone detection task. On the other hand, the Vision Transformer (ViT) [3] represents a novel application of the Transformer model, which has become a favored choice for various natural language processing (NLP) tasks including machine translation, question answering, text classification, document summarization, and more [13]. A crucial aspect of the Transformer's success lies in its capacity to comprehend intricate interdependencies among long input sequences through self-attention [10]. With the introduction of the Vision Transformer (ViT), it has been demonstrated for the first time that the transformer architecture can be applied directly to image-based tasks. This is accomplished by perceiving an image as a series of patches and feeding these patches into an encoder network based on multi-headed self-attention layers [3]. In our drone dataset, we only use a plain ViT-b16 model plus a few top layers to prove that it can achieve much higher accuracy than leading-edge convolutional networks VGG 16 and ResNet 50 at a cost of longer training time and larger model. To further compare the CNN and ViT based networks on more complex object detection challenge, we experiment with the only open-source ViT-based YOLO detector, You Only Look At One Sequence (YOLOS), which uses a ViT as backbone and a simplified detector head without any performance-oriented designs [4]. By doing so, we demonstrate that while ViT can more efficiently capture long-range dependencies between image patches via self-attention, it also requires more training data and a careful design to perform well. Finally, we analyze and summarize the distinct features of CNN and ViT-based models, and their unique advantages in different scenarios.

## II. Dataset

The drone dataset [6] consists of 1359 images, 1297 of which contain only a single drone, while the remaining 62 images feature multiple drones. Along with these images, the dataset also provides the X and Y coordinates for the drone's center in each image, as well as the height and width necessary to draw bounding boxes for object detection (Figure 1). Currently, due to the limited number of multi-object training images and for the sake of simplicity, we are only using the images containing a single drone for training.

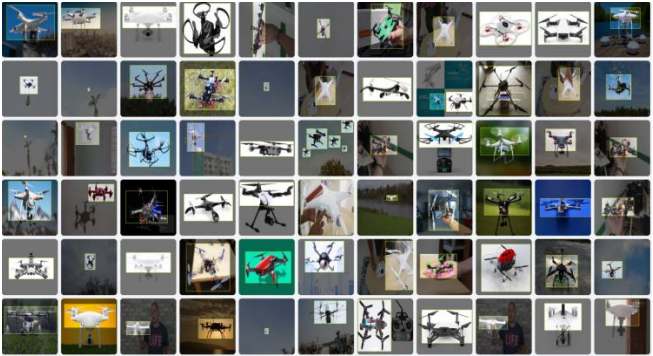

**Figure 1.** Drone dataset visualization

Analyzing the filtered dataset's complexity, we look at the X and Y coordinates of the center of the drone relative to the image for spatial complexity and the area ratio is defined as Equation 1 for perspective complexity. Using a 3D visualization and bubble chart, we are able to better understand and visualize the drone dataset (Figure 2-3). From the graphs, there appears to be a Gaussian distribution across spatial and perspective complexity in the drone dataset, showing that the filtered drone dataset images are not heavily skewed or biased by the Central Limit Theorem. The visualization also shows that a large portion of the drones in the dataset are very small and therefore hard to be located accurately.

$$area\ ratio = \frac{size\ of\ bounding\ box\ (px)}{size\ of\ image\ (px)} \qquad Equation\ (1)$$

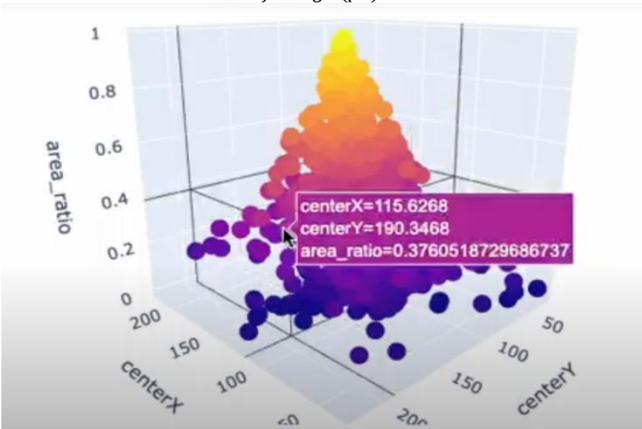

**Figure 2.** 3D visualization of Spatial and Perspective Complexity on Drone Dataset

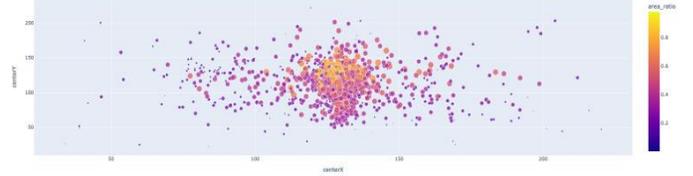

**Figure 3.** Bubble chart of drone bounding box size (area ratio) and coordinates (centerX and centerY)

To test ViT's full capability of multi-object detection, we further augment the dataset by applying mosaic, noise, flip, rotation and blur, so that the training dataset size increases to 3021 and there are more images containing multiple drones (Figure 4). We have trained YOLO v7, YOLO v8, and YOLOS on the augmented dataset.

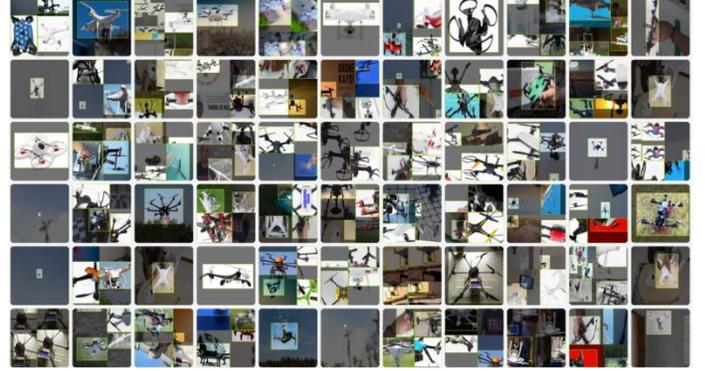

**Figure 4.** Augmented drone dataset visualization

## III. Methods and Implementation

### A. Single-Drone Detection

A 14 layers vanilla customized CNN model (Figure 5) and three popular transfer learning models are used as backbones to tackle the drone detection challenge: ResNet50 (Figure 6), VGG16 (Figure 7) and ViT-b16 (Figure 8). As it is shown in the following figures, we design the same top 5 layers for them for better comparison. Since it is a regression task, the MSE loss function and linear activation function are used on the final output layer. The Adam optimizer is deployed as usual with a learning rate of 0.0001. Notably, all the images are reshaped to (256,256,3) before training to fit in the vision transformer's unique patch size (16 by 16). Compared to classical CNN, ResNet and VGG network, the ViT-b16 transformer is more computationally intensive. With our ViT-b16 model having 86 million trainable parameters (Figure 8), our best performing CNN model VGG16 only has 19 million trainable parameters (Figure 7).

```
Layer (type)                 Output Shape              Param #
=================================================================
conv2d_6 (Conv2D)            (None, 256, 256, 32)      896
average_pooling2d_6 (Averag  (None, 128, 128, 32)      0
ePooling2D)
conv2d_7 (Conv2D)            (None, 128, 128, 64)      18496
average_pooling2d_7 (Averag  (None, 64, 64, 64)        0
ePooling2D)
conv2d_8 (Conv2D)            (None, 64, 64, 128)       73856
average_pooling2d_8 (Averag  (None, 32, 32, 128)       0
ePooling2D)
flatten_2 (Flatten)          (None, 131072)            0
dense_8 (Dense)              (None, 128)               16777344
dropout_6 (Dropout)          (None, 128)               0
dense_9 (Dense)              (None, 64)                8256
dropout_7 (Dropout)          (None, 64)                0
dense_10 (Dense)             (None, 32)                2080
dropout_8 (Dropout)          (None, 32)                0
dense_11 (Dense)             (None, 4)                 132
=================================================================
Total params: 16,881,060
Trainable params: 16,881,060
Non-trainable params: 0
```

**Figure 5.** Vanilla CNN model summary

```
conv5_block3_3_conv (Conv2D)    (None, 8, 8, 2048)   1050624  ['conv5_block3_2_relu[0][0]']
conv5_block3_3_bn (BatchNormal  (None, 8, 8, 2048)   8192     ['conv5_block3_3_conv[0][0]']
ization)
conv5_block3_add (Add)          (None, 8, 8, 2048)   0        ['conv5_block2_out[0][0]',
                                                                'conv5_block3_3_bn[0][0]']
conv5_block3_out (Activation)   (None, 8, 8, 2048)   0        ['conv5_block3_add[0][0]']
flatten (Flatten)               (None, 131072)       0        ['conv5_block3_out[0][0]']
dense (Dense)                   (None, 128)          16777344 ['flatten[0][0]']
dense_1 (Dense)                 (None, 64)           8256     ['dense[0][0]']
dense_2 (Dense)                 (None, 32)           2080     ['dense_1[0][0]']
dense_3 (Dense)                 (None, 4)            132      ['dense_2[0][0]']
=================================================================
Total params: 40,375,524
Trainable params: 40,322,404
Non-trainable params: 53,120
```

**Figure 6.** ResNet 50 model summary

```
block5_pool (MaxPooling2D)   (None, 8, 8, 512)         0
flatten_1 (Flatten)          (None, 32768)             0
dense_4 (Dense)              (None, 128)               4194432
dense_5 (Dense)              (None, 64)                8256
dense_6 (Dense)              (None, 32)                2080
dense_7 (Dense)              (None, 4)                 132
=================================================================
Total params: 18,919,588
Trainable params: 18,919,588
Non-trainable params: 0
```

**Figure 7.** VGG16 model summary

```
Layer (type)                 Output Shape              Param #
=================================================================
vit-b16 (Functional)         (None, 768)               85844736
flatten_3 (Flatten)          (None, 768)               0
dense_6 (Dense)              (None, 128)               98432
dense_7 (Dense)              (None, 64)                8256
dense_8 (Dense)              (None, 32)                2080
dense_9 (Dense)              (None, 4)                 132
=================================================================
Total params: 85,953,636
Trainable params: 85,953,636
Non-trainable params: 0
```

**Figure 8.** ViT-b16 model summary

Before analyzing the vision transformer, we need to first understand the working principles of the ViT-b16 model. An image is split into fixed-size patches (Figure 9), then linearly embedding and position embeddings (Figure 10) are added to each of them. The necessity of the positional embeddings comes from the invariance of output context vector with respect to different permutation of input patches. In a self-attention layer, each output vector is calculated as a weighted sum of all value vectors (Figure 11), and the weights (i.e., attention scores) are based solely on the content of the key and query vectors and not their order (Figure 12). As a result, despite the output context vector order will be permuted in the same way as we permute the input sequence, the set content of output context vectors will remain the same. Then, the resulting sequence of vectors from embedding layers are fed to a standard Transformer encoder (Figure 9), which contains six transformer layers in our case. One transformer layer consists of one self-attention layer followed by a dense layer. Different from a CNN layer which only captures local context information by sliding a small kernel window, a self-attention layer treats all image patches equally, so it is inherently better at detecting long-range dependency between patches and requires a larger dataset to efficiently learn these dependencies. In order to perform classification, we use the standard approach of adding an extra learnable "classification token" to the sequence [12]. Unlike other transformers, the vision transformer does not have a decoder network, allowing our model to improve its training efficiency significantly, since we only need to train the embedding layers, a transformer encoder network, and a SoftMax classifier. In our case, we use 5 layers MLP with a bounding box detector head instead of a SoftMax classifier.

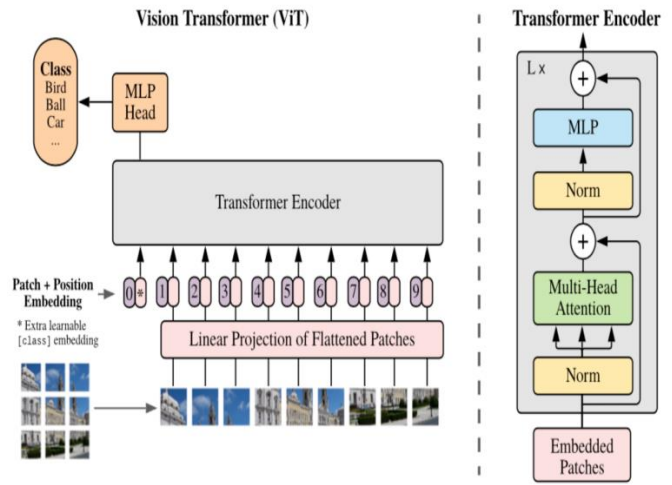

**Figure 9.** Vision Transformer model architecture

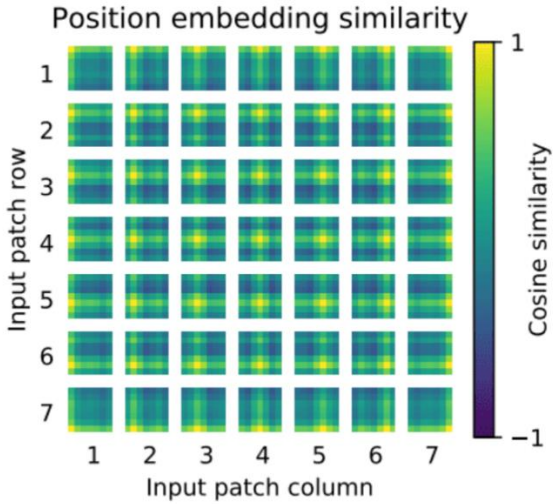

**Figure 10.** Positional embedding visualization

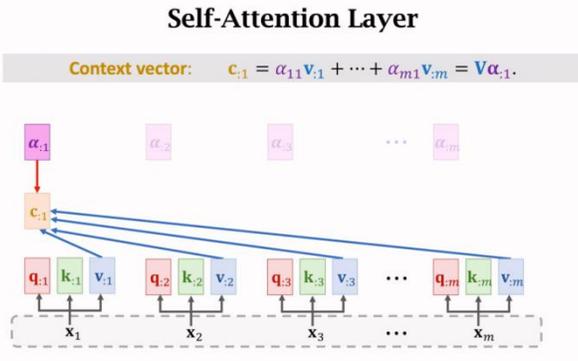

**Figure 11.** Output context vector calculation in the self-attention layer

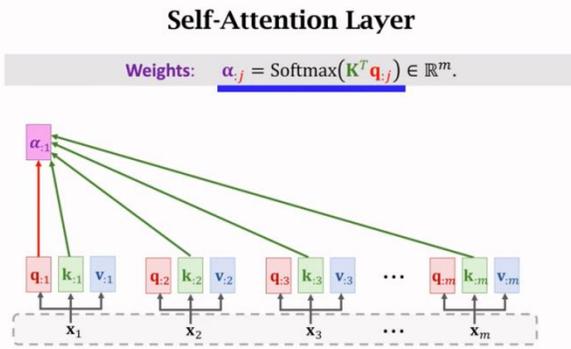

**Figure 12.** Attention scores calculation in the self-attention layer

### B. Multi-Drone Detection

YOLO v7, YOLO v8 and YOLOS are trained on the augmented multi-drone dataset. Unlike the YOLO models (Figure 13) which use CNNs as backbones and complex heads, YOLOS uses ViT as backbone (Figure 14) and only an MLP of 2 hidden layers to implement both classification and bounding box regression heads [4]. In essence, YOLOS is not optimized for better performance but to prove the transferability of ViT [4]. The transformation from a Vision Transformer (ViT) to a YOLOS detector is straightforward. First, YOLOS eliminates the [CLS] token used for image classification and instead adds a hundred learnable detection tokens ([DET] tokens), which are randomly initialized, to the input patch embeddings ([PATCH] tokens) for the purpose of object detection. Secondly, during the training process, YOLOS substitutes the image classification loss found in ViT with a bipartite matching loss, enabling it to carry out object detection in a set prediction manner, in line with the method proposed by DETR [2], the first attempt to successfully apply transformers to multi-object detection.

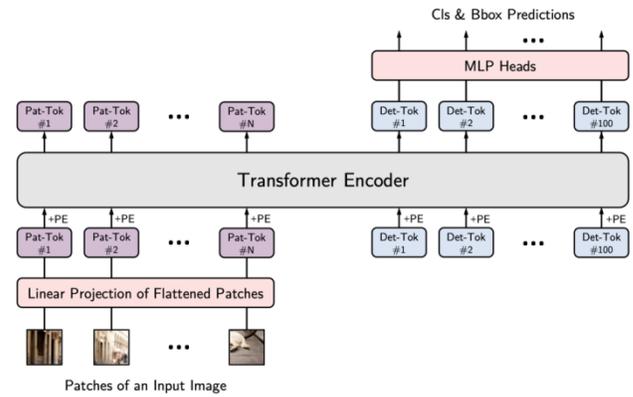

**Figure 13.** YOLO v8 model summary

**Figure 14.** YOLOS model architecture

### IV. Results and Analysis

#### A. Single-Drone dataset

For the sake of simplicity, we present only the plots for ViT-b16 and our highest-performing CNN transfer learning model: VGG16. As illustrated in the plots below, we successfully achieved 89% and 95.3% validation accuracy on our drone dataset using our VGG16 and ViT-b16 models, respectively, after just 50 epochs. When comparing the two models, the ViT-b16 transformer loss is 4.6 times lower than our best model, VGG16, with the VGG16 loss standing at around 0.0067 and the transformer loss approximately 0.00145 (Figure 15). For reference, we also plot the IoU comparison (Figure 15). It is clear that the CNN model lacks the capacity to capture deeper data dependencies and begins to overfit the drone dataset early on. In contrast, our plain ViT model does not demonstrate an apparent overfitting pattern and manages to attain higher accuracy in the validation dataset. This highlights the power and potential of the vision transformer in object detection tasks through the self-attention mechanism (Figure 16).

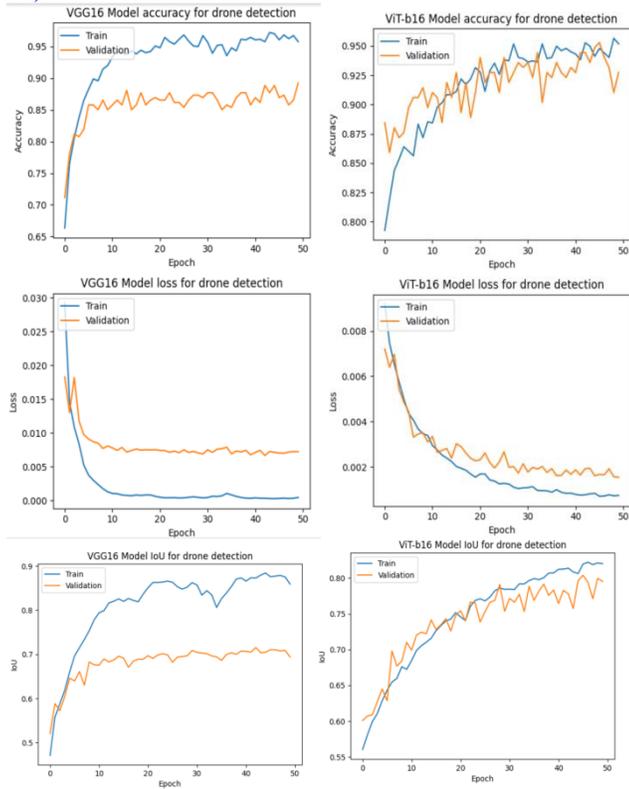

**Figure 15.** Comparison between VGG16 and Vit-b16 on validation dataset

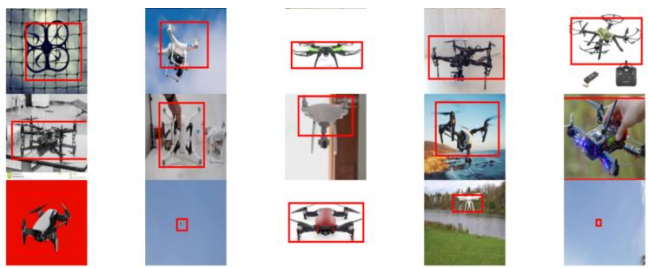

**Figure 16.** Vit-b16 bounding box prediction visualization

It's important to note that while ViT requires 1s/step for training, VGG16 only requires 0.33s/step, meaning the training speed of ViT is nearly three times slower than that of VGG16. Other CNN models are even faster than VGG16, but their accuracy is not comparable. The cost of achieving superior accuracy (referring to validation loss here) is increased computational power, longer training time, and the use of a more complex structure. However, in the crucial task of fast-moving drone detection, and with the support of rapidly advancing GPUs, we prioritize detection accuracy, and thus, this tradeoff is completely acceptable.

### B. Multi-Drone dataset

All the results from our drone dataset seem satisfactory now, but we are more curious about ViT's performance on complex object detection tasks such as multi-drone detection. For simplicity purpose here we only compare results of YOLOS and our best performing YOLO model: YOLO v7. With the help of Roboflow online training, the result of YOLO v7 is amazing (Figure 18). It reaches around 98% mAP after 200 epochs, indicating almost a perfect performance (Figure 17). On the other hand, the result of YOLOS is also inspiring (Figure 20). After only 20 epochs, mAP reaches around 96% (Figure 19), while YOLO v7 mAP is only around 89% at the 20[th] epoch. This again demonstrates ViT-based model's promising ability in larger scale object detection.

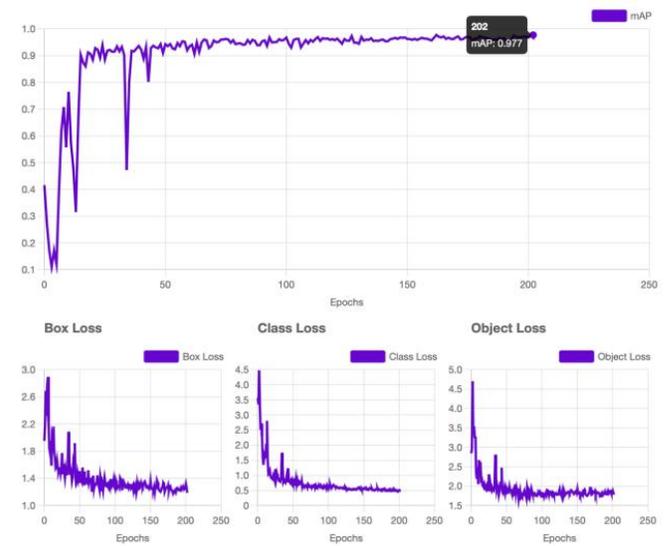

**Figure 17.** YOLO v7 performance over 200 epochs

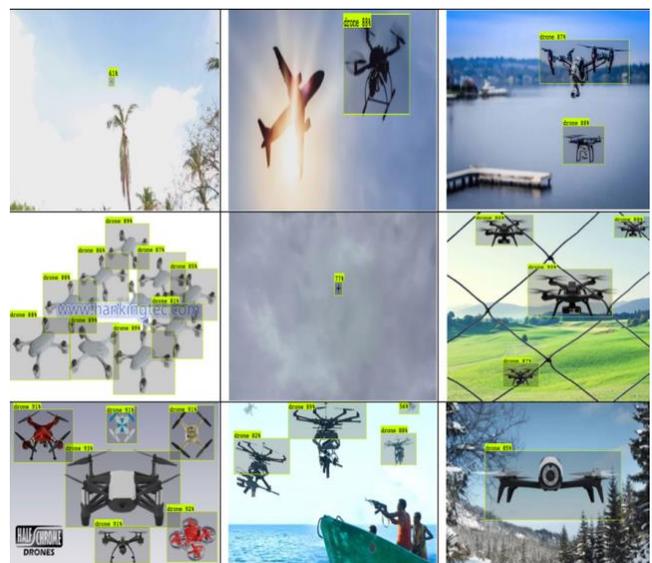

**Figure 18.** YOLO v7 bounding box prediction visualization

```
IoU metric: bbox
 Average Precision  (AP) @[ IoU=0.50:0.95 | area=   all | maxDets=100 ] = 0.633
 Average Precision  (AP) @[ IoU=0.50      | area=   all | maxDets=100 ] = 0.958
 Average Precision  (AP) @[ IoU=0.75      | area=   all | maxDets=100 ] = 0.706
 Average Precision  (AP) @[ IoU=0.50:0.95 | area= small | maxDets=100 ] = 0.127
 Average Precision  (AP) @[ IoU=0.50:0.95 | area=medium | maxDets=100 ] = 0.410
 Average Precision  (AP) @[ IoU=0.50:0.95 | area= large | maxDets=100 ] = 0.675
```

**Figure 19.** YOLOS mAP (same as AP here) performance over 20 epochs on the augmented dataset

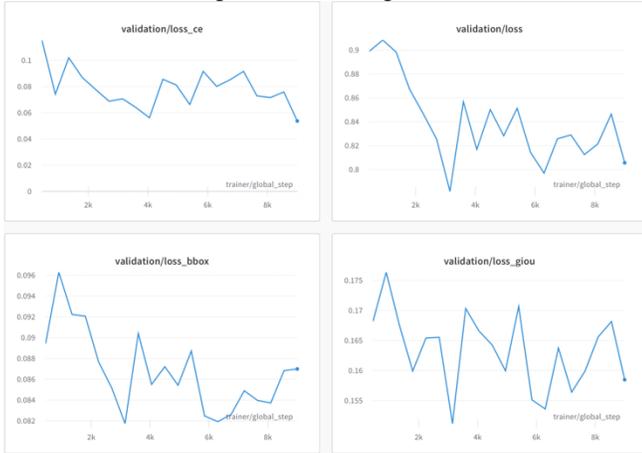

**Figure 20.** YOLOS validation loss performance over 20 epochs on the augmented dataset

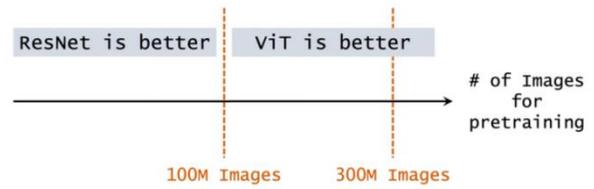

**Figure 21.** ViT's classification performance beats CNN on larger datasets

```
IoU metric: bbox
 Average Precision  (AP) @[ IoU=0.50:0.95 | area=   all | maxDets=100 ] = 0.617
 Average Precision  (AP) @[ IoU=0.50      | area=   all | maxDets=100 ] = 0.924
 Average Precision  (AP) @[ IoU=0.75      | area=   all | maxDets=100 ] = 0.726
 Average Precision  (AP) @[ IoU=0.50:0.95 | area= small | maxDets=100 ] = 0.032
 Average Precision  (AP) @[ IoU=0.50:0.95 | area=medium | maxDets=100 ] = 0.332
 Average Precision  (AP) @[ IoU=0.50:0.95 | area= large | maxDets=100 ] = 0.653
```

**Figure 22.** YOLOS mAP (same as AP here) performance over 20 epochs on the original dataset

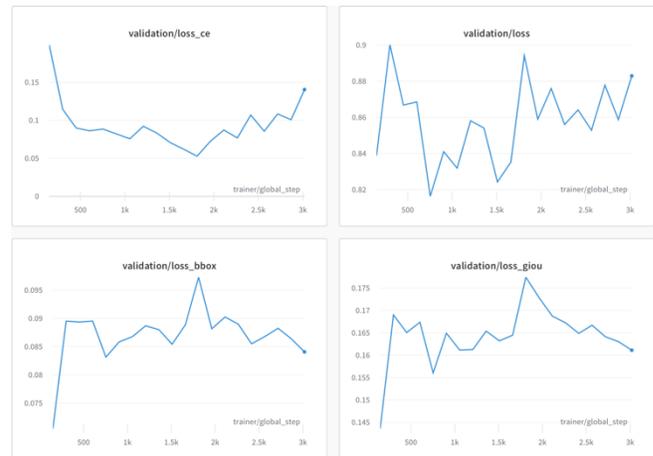

**Figure 23.** YOLOS validation loss performance over 20 epochs on the original dataset

However, similar to ViT-b16, YOLOS suffers from the same problem of requiring longer training time and a larger model. Specifically, YOLOS-small contains 30.7 million trainable parameters, while YOLOv8-small only contains 11 million (Figure 13). Every YOLOS training epoch takes around 10~12 minutes, which is more than 15 times longer than the performance-oriented YOLO v8 and YOLO v7, so we only utilize the first 20 epochs for demonstration purpose.

Other research reveals another key advantage of vision transformers in object classification over CNNs. When the training dataset size is sufficiently large (more than 100 million), ViT outperforms CNN by larger margins [3] (Figure 21). This is intuitively understandable, given ViT's more complex model structure. It is also worthwhile to validate this in the object detection task, so we train our YOLOS model on the original dataset of only 1359 images without any data augmentation, with only 62 of them containing multiple drones in one image. We found that the mAP drops quickly from 96% to around 92% (Figure 22), performing much closer to YOLO models (89%). More importantly, the validation loss plots on the original dataset appear much worse and unstable (Figure 23), suggesting that the model might not even be learning at all. From this, we conclude that the dataset size is extremely important for ViT-based models in object detection tasks.

Worth mentioning, the experiment on YOLOS is just a starting point for comparing ViT and CNN, since YOLOS does not specifically adopt YOLO's architecture, such as darknet, decoupled head, DFL loss and so on. More accurate comparison may be conducted between the YOLO v8 and ViT-YOLO [13], which aligns better with YOLO structure and ranks 2nd globally in the VisDrone-DET2021 competition [1]. Unfortunately, this model is not open-source, but we still believe this ViT&YOLO hybrid model can easily surpass YOLO models based on its outstanding performance on the global challenge. Additionally, ViT models also prove great potential in multi-label object classification task, as ADDS (ViT-L-336) is currently ranked 1st globally in the Microsoft COCO multi-label classification challenge [7].

## V. Conclusion and Discussion

In the comparative analysis conducted, we identified four main insights about the performance of Vision Transformer (ViT) and Convolutional Neural Network (CNN) based deep learning models in the realm of object detection. Primarily, ViT-based models consistently demonstrate superior performance over CNN-based models in similar epochs for object detection and classification tasks. This can be ascribed to the built-in self-attention mechanism in ViT that effectively captures long-range data dependencies. However, the use of ViT networks brings with it the demand for increased computational power and a lengthier training period, a stark contrast to the requirements of CNN networks. Furthermore, the extent of training data plays a critical role in ViT's commendable performance. A combination of large-scale training datasets and data augmentation methodologies ensures ViT's dominance over CNN. Finally, it should be noted that our study is centered solely around pure ViT models. For ViT to surpass the performance of leading-edge CNN models in more complex tasks, a more intricately designed structure would be necessary to elevate ViT's performance further.

## VI. Acknowledgement

The author would like to thank Tyler Yu from University of California, Irvine Computer Science department for creating the 3D visualization and bubble chart for the drone dataset.